\documentclass[sigchi]{acmart}

\usepackage[capitalise, noabbrev]{cleveref}
\usepackage{multirow}
\usepackage{xcolor, soul}
\sethlcolor{green}

\AtBeginDocument{%
  }

\setcopyright{acmcopyright}
\copyrightyear{2022}
\acmYear{2022}
\acmDOI{10.1145/3544794.3558464}

\acmConference[ISWC '22]{ISWC '22: ACM International Symposium on Wearable Computers}{September 11--15, 2022}{Atlanta, USA and Cambridge, UK}
\acmBooktitle{ISWC '22: ACM International Symposium on Wearable Computers,
  September 11--15, 2022, Atlanta, UK and Cambridge, UK}

\acmSubmissionID{1095}

\begin{document}

\title[Learning from the Best: Contrastive Representations Learning Across Sensor Locations]{Learning from the Best: Contrastive Representations Learning Across Sensor Locations for Wearable Activity Recognition}

\author{Vitor Fortes Rey, Sungho Suh, Paul Lukowicz}
\affiliation{%
  \institution{German Reserch Center for Artificial Intelligence (DFKI) and University of Kaiserslautern, Germany}  
  \country{}
  \streetaddress{Trippstadterstr. 122}
  \postcode{67663}
}
\email{{vitor.fortes_rey, sungho.suh, paul.lukowicz}@dfki.de}

\begin{abstract}
  We address the well-known wearable activity recognition problem of having to work with sensors that are non-optimal in terms of information they provide but have to be used due to wearability/usability concerns (e.g. the need to work with wrist-worn IMUs because they are embedded in most smart watches). To mitigate this problem we propose a method that facilitates the use of information from sensors that are only present during the training process and are unavailable during the later use of the system. The method transfers information from the source sensors to the latent representation of the target sensor data through contrastive loss that is combined with the classification loss during joint training (\cref{fig:teaser}). 
  We evaluate the method on the well-known PAMAP2 and Opportunity benchmarks for different combinations of source and target sensors showing average (over all activities) F1 score improvements of between 5\% and 13\% with the improvement on individual activities, particularly well suited to benefit from the additional information going up to between  20\% and 40\%. 
\end{abstract}


\begin{teaserfigure}
  \centering
  \includegraphics[width=0.91\textwidth]{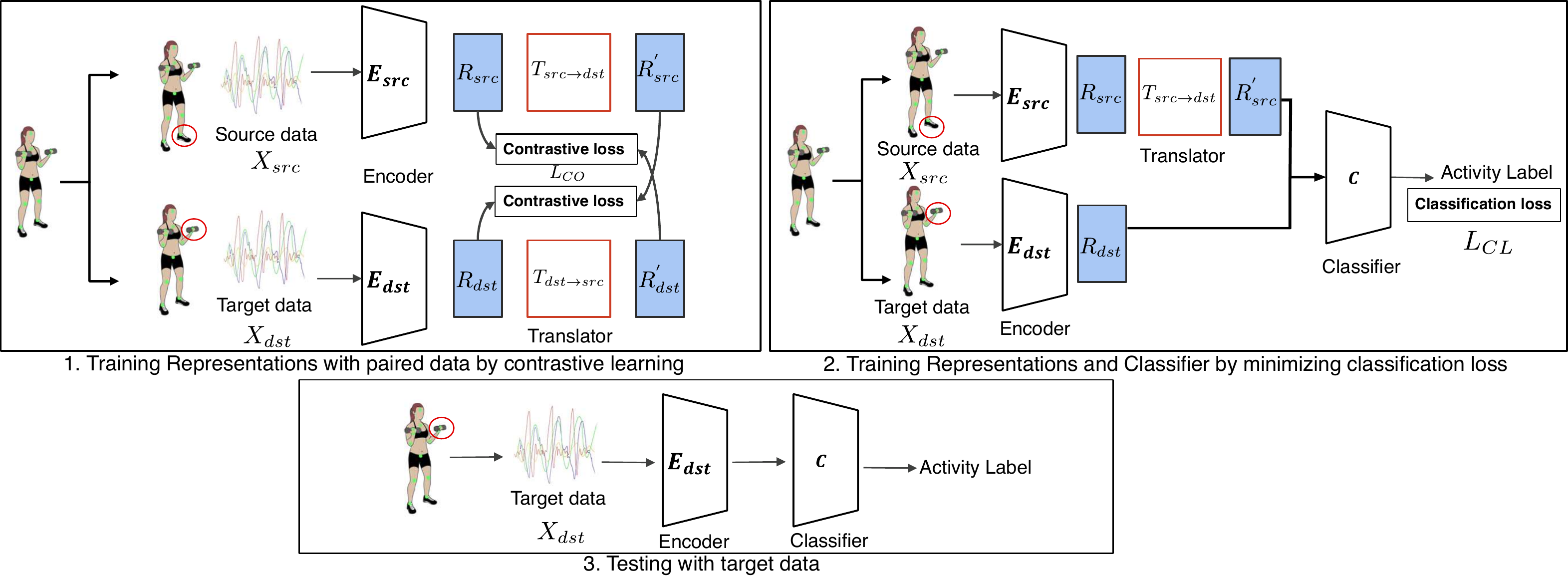}
  \caption{Overall network architecture and steps of the proposed method.}
  \label{fig:teaser}
\end{teaserfigure}

\maketitle

\section{Introduction}
A well-known problem of wearable human activity recognition (HAR) is the fact that sensor locations that are feasible for long-term, everyday deployment often provide poor and/or noisy information for the recognition of activities that may be relevant for a particular application. Thus, for example, with the widespread adoption of smart watches wrist-worn, IMUs are a broadly available, unobtrusive sensing modality suitable for everyday use. However, it is well-known that many wearable HAR tasks such as wrist-worn IMUs are a poor source of information. For example, when looking at a detailed analysis of modes of locomotion (walking up, down, running, etc.) most/best information is contained in the motion of the legs and the motion patterns of the trunk. Arms on the other hand are often moved independently of the modes of locomotion (e.g. gesticulating while talking while walking). Locomotion-related signals do propagate to the wrist but since they propagate through several joints they are significantly weakened and often overshadowed by signals related to other arm activities.  

We propose a method to mitigate this problem by leveraging contrastive representation learning to transfer knowledge between sensors on different body locations. The proposed method is based on the following observations and ideas:
\begin{enumerate}
    \item Following the general trend in ML recently HAR methods that first map the sensor data onto a latent representation and then perform the classification have become increasingly popular. It was recently shown that such a method based on a spatio-temporal transformer outperforms the state-of-the-art approaches on standard benchmarks \cite{chavarriaga2013opportunity, reiss2012introducing, zhang2022if}.
    \item It has been shown that domain knowledge can be encoded into such latent representations independently of the actual classification task (so-called downstream task). Thus, for example, state-of-the-art Self Supervised Learning approaches \cite{devlin2018bert,morgado2020learning, morgado2021audio,he2022masked} (pre) train the latent representation on so-called pretext tasks that are related to but not identical to the downstream task for which the system is actually intended. 
    \item System training (or at least parts of it) is often performed in constrained environments where it is often easily possible to include a broad variety of sensors, even if some of them may not be suitable for long-term real-world use.
\end{enumerate}

From the above considerations, we propose to train the system not just with the sensor that will later be deployed (target sensor)  but also with additional sensors that may contain useful information (source sensors). The information from the source sensors is transferred to the latent representation of the target sensor through contrastive loss that is combined with the classification loss during joint training (\cref{fig:teaser}). The contrastive loss forces the structure of the target sensor representation (relative position of activities in the space) to reflect what the source sensor knows about the relationship between the activities to be recognized. The classification loss ensures that the target sensors map the signals to the correct regions of the latent space.  

Our method can also be applied to real-world applications going beyond IMU sensors since it is suited to long-term use, for example, applied to human-robot collaboration in manufacturing lines that require HAR to improve productivity and quality assurance and to keep workers safe in human-robot collaboration.

\section{Related Work}
HAR has emerged as an important research problem in the field of signal processing and computer vision. Over the past few years, deep learning methods have emerged as the state-of-the-art approaches in HAR by utilizing convolutional neural networks (CNN) and long short-term memory networks (LSTM) \cite{yang2015deep, ordonez2016deep, carreira2017quo, gupta2021quo, suh2022adversarial}. 

Recently, other methods that also improve HAR by taking advantage of modalities available only at training time have been developed \cite{yang2022more, lago2021using}. M2L \cite{yang2022more} proposes as system that learns using more modalities than available during testing by knowledge transfer using the cosine similarity between feature representations. The direction of transfer is controlled by transferring only when one model exceeds the other’s performance. On the other hand, it cannot yet use unlabelled data and is still limited by the use of hand-crafted features. In comparison, our method has a single classifier for all modalities, as they are converted to the same representation. Moreover, it is not limited by hand-crafted features and can use unlabelled data.
Another work \cite{lago2021using} proposes a method where clustering and boosting are used to improve single-sensor performance when more sensors are available at training time. It does it by performing clustering with all features and then mapping the single-sensor representation to cluster values in the all-sensors one using multi-valued regression. It can use unlabelled data in the clustering step and later labelled data to classify activities using clusters-as-features. Our method can also use unlabelled data, but does not use clustering and instead relies on a contrastive learning framework. Another important distinction is  that ours unifies the other sensor's and the single-sensor’s representation by translating between the two in order to improve the latter, while theirs brings the single-sensor data into a joint unsupervised representation. In their case one needs to choose the representation space size so that the single-sensor can still be correctly mapped to a meaningful cluster in the shared representation. Moreover, it is not clear how their method deals with limitations present in this mapping as well as the possible loss of relevant feature information by clustering.

\section{Proposed Method}\label{sec:method}
		The goal of HAR is to predict the activity that is being performed, out of a set of possible activities $N_a$, for a specific time interval, usually a sliding window of size $n_w$, and data channel $c$. 
		We consider  two synchronized sensor data streams (modalities):  source $X_{src}=[x_{src}^1, x_{src}^2, ...]$ and destination  $X_{dst}=[x_{dst}^1, x_{dst}^2, ...]$, where the source data $x_{src,i} \in \mathbb{R}^{c_{src} \times n_w}$ is only present in training, while the destination data $x_{dst,i} \in \mathbb{R}^{c_{dstI} \times n_w}$ is present both in training and deployment. 
		The aim is to provide a better classification with $Y=[y^1, y^2, ...]$ the corresponding activity label set of $X$. 
		As shown in \cref{fig:teaser}, the proposed framework is composed of three components: encoder, translator, and classifier. 
	   There is a dedicated encoder for each modality ($E_{src}: X_{src} \rightarrow R_{src}$ and $E_{dst}: X_{dst} \rightarrow R_{dst}$) to map the sensor data onto a respective  spatiotemporal latent representations $R_{src}$ ($R_{dst}$). 
	   To relate the source and the target latent representations to each other there are two translator networks:  $T_{src \rightarrow dst}$ is the network that translates $R_{src}$ to $R_{dst}$, with $T_{dst \xrightarrow{} src}$ doing the same in reverse.  For classification we train only a single network $C: R_{shrd} \rightarrow Y$ where $R_{shrd}$ is either $R_{dst}$ or $R^{'}_{src}=T_{src \rightarrow dst}(R_{src})$ (Fig {\ref{fig:teaser}} right). We call this $R_{shrd}$ our shared representation as it represents the target's latent representation (which will be used during deployment) coming from either source. 
	   
	   For training we consider two types of loss: (1) the classification loss and (2) the contrastive representation misalignment loss between the source and target modality.  The classification loss is computed by taking every sample, feeding each modality to the corresponding encoder and then to the classifier (with the output of the source encoder being translated to the target representation before being input to the classifier) as shown in Fig \ref{fig:teaser} on the right. In detail the classification loss is expressed as
			\footnotesize
			\begin{equation}
				L_{CL} = \sum_i { L_{CE}(
					C( E_{dst}(x_{dst}^{i})),
					y^{i} 
					) 
					+ L_{CE}(
					C(T_{src \rightarrow dst} ( E_{src}(x_{src}^{i}) )),
					y^{i}
					)
				}
			\end{equation}
			\normalsize
			where $L_{CE}$ is the categorical cross entropy loss, defined as 
			\begin{equation}
				\label{eq:ce}
				L_{CE}(\tilde{z},z) = - z \log (\tilde{z})
			\end{equation}
	   and the classification path for a point $x_{dst}^i$  being $C( E_{tgt}(x_{dst}^i) )$, while for a point in $x_{src}^i$ it is $C(T_{src \rightarrow dst} ( E_{src}(x_{src}^i) ))$.
	   
	   The representation misalignment loss is the instrument that allows us to transfer knowledge from the source to the target modality. To compute it for each combination of data items in a  batch one is input to the source (using the source modality)  and one to the target (using the target modality) encoder. Every batch consists of $N$ randomly sampled $\big\{x_{src}^{i}, x_{dst}^{i}\big\}^{N}_{i=0}$, where $x_{src}^{i}$ occurs at the same time as $x_{dst}^{i}$. The contrastive loss $L_{CO}$ is computed with the help of the translator networks as indicated in \ref{fig:teaser} on the left and penalizes the misalignment of the representations output by the two encoders \emph{if the items do not belong to the same time interval}. It thus forces both encoders to have a similar representation of the same point in time effectively transferring knowledge between the two corresponding sensor modalities (even if class labels are not available). 
	   Specifically, the contrastive loss is defined as follows:
	        \footnotesize
			\begin{equation}
				\begin{split}
					L_{CO} = \sum_i \sum_k \Big(
					& L_{N}(
					E_{dst}(x_{dst}^{k}),
					T_{src \rightarrow dst}(E_{src}(x_{src}^{k})),
					\big\{ E_{dst}(x_{dst}^{i}) \big\}^{N}_{i=0}
					) \\
					&+ L_{N}(
					E_{src}(x_{src}^{k}),
					T_{dst \rightarrow src}(E_{dst}(x_{dst}^{k})),
					\big\{E_{src}(x_{src}^{i}) \big\}^{N}_{i=0}
					) \Big)    
				\end{split}
			\end{equation}
			\normalsize 
			
			where $L_{N}$ is the infoNCE loss \cite{van2018representation}, defined as
			\small
			\begin{equation}
				L_{N}(x,x_t, P_x) = -\log\frac{exp( cosSim(x_t, x)) / \tau}{ \sum_{x_p \in P_x}{ exp(cosSim(x_p, x)) / \tau}}
			\end{equation}
			\normalsize 
			where $\tau$ is a temperature hyper-parameter and $cosSim$ is the cosine similarity between $x$ and $x_t$ in a smaller space (in our case averaging across time).
	   

		
\begin{figure}
			\centering
			\includegraphics[width=0.99\columnwidth]{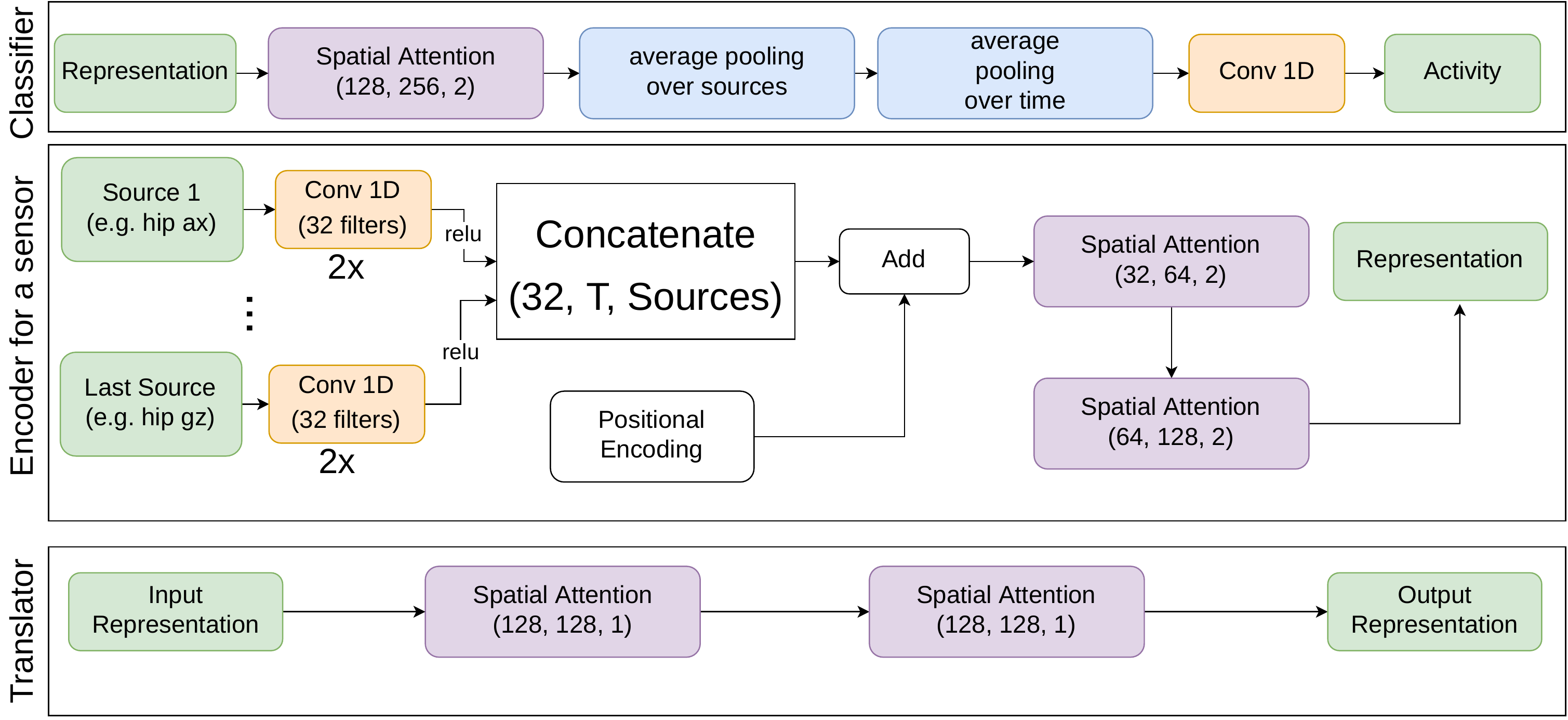}
			\caption{Network architectures of the proposed spatial-temporal transformer (ST-Transformer) composed of encoder, translator, and classifier.}
			\label{fig:all_nets}
		\end{figure}

		The encoder, classifier, and translator networks are all based on a Transformer architecture. 
		Recently, Transformers have obtained state-of-the-art results in many fields, including HAR \cite{plizzari2021skeleton}. They are also the architecture of choice for most cross-modal contrastive learning approaches \cite{morgado2020learning}. The specific transformer concept used in this work is based on the spatial transformer  \cite{plizzari2021skeleton} where attention is performed between the components (skeleton joints in \cite{plizzari2021skeleton} and any meaningful separation inside a modality such as individual acceleration and gyro channels of the IMU in this work). In this paper, we propose a Spatio-temporal transformer (ST-Transformer) as shown in \cref{fig:all_nets} and apply it to standard HAR benchmarks also used in this work (see table \ref{table:res2}) to achieve results superior to the state-of-the-art methods. As a consequence, we have decided to use it as a basis for this work, although as described above we also investigate a CNN-based encoder for comparison and to demonstrate that the general concept of contrastive representation learning across sensor modalities is not restricted to a specific encoder architecture.

			
			
		
\section{Evaluation}
\label{sec:experimentalresults}

    \begin{table*}[!t]
    \centering
    \caption{Results for learning between different sensor positions in the PAMAP2 dataset. Leaving one user out.}
    \label{table:resPosEncodingDrillBetweenPositions}
    \begin{tabular}{llllllll}
    \toprule
    mode & teacher & target &     macro F1          &       improvement               &       best                   &        2nd best                  &       worst                \\
    \midrule
    CFSR & ANKLE & CHEST &  0.70 $\pm$ 0.11 &         0.08  &     Nordic walking:0.28  &        watching TV:0.19  &           lying:-0.0  \\
                       &      & HAND &  0.62 $\pm$ 0.15 &         0.08 &  descending stairs:0.40  &            walking:0.20  &         sitting:-0.0  \\
                       & CHEST & ANKLE &  0.53 $\pm$ 0.18 &         0.08  &   ascending stairs:0.28  &        watching TV:0.22  &         running:-0.1  \\
                       &      & HAND &  0.62 $\pm$ 0.16 &         0.13  &            running:0.26  &            walking:0.25  &        standing:0.04  \\
                       & HAND & ANKLE &  0.57 $\pm$ 0.19 &         0.12  &              lying:0.26  &      computer work:0.20  &         sitting:-0.0  \\
                       &      & CHEST &  0.70 $\pm$ 0.10 &         0.10  &            running:0.26  &              lying:0.19  &        standing:-0.0  \\
    \midrule
    \midrule
    Baseline & - & ANKLE &  0.45 $\pm$ 0.16 &   CHEST &   0.61 $\pm$ 0.12  &    HAND  &            0.51 $\pm$ 0.13  \\
    \bottomrule
    \end{tabular}
    \end{table*}
    \begin{table*}[!t]
    \centering
    \caption{Results for leaving one user out in the PAMAP2 dataset (all sensor combinations) and Opportunity dataset (only Rwrist as target). Baselines with all sensors are ours ST-Transformer, MC-CNN \cite{yang2015deep}, DeepConvLSTM \cite{ordonez2016deep}, and Self-Attention \cite{mahmud2020human}. }
    \label{table:res2}
    \resizebox{\linewidth}{!}{
    \begin{tabular}{llllc}
    \toprule
     dataset  & mode &  macro F1  &  accuracy             &  \small improvement\\
    \midrule
    \multirow{4}{*}{PAMAP2}  & CFSR               &  0.63 $\pm$ 0.16 &  0.70 $\pm$ 0.16 &     0.11  \\
            & CTSR               &  0.58 $\pm$ 0.16 &  0.66 $\pm$ 0.17 &     0.06  \\
            & TSR                 &  0.61 $\pm$ 0.16 &  0.68 $\pm$ 0.15 &     0.08  \\
            & only target sensor &  0.52 $\pm$ 0.16 &  0.60 $\pm$ 0.15 &           n/a \\
    
    \midrule
    
    \multirow{2}{*}{Baseline all sensors}  & MC-CNN / DeepConvLSTM&  0.68 $\pm$ 0.16 / 0.57 $\pm$ 0.17  & 0.76 $\pm$ 0.16 / 0.66 $\pm$ 0.19 \\
                &Self-Attention / \textbf{ST-Transformer}& 0.69 $\pm$ 0.15 / \textbf{ 0.73 $\pm$ 0.13} & 0.77 $\pm$ 0.15 / \textbf{ 0.81 $\pm$ 0.13} \\

    \midrule
    
    
    
    
    \multirow{2}{*}{Opportunity}               & CFSR               &  0.30 $\pm$ 0.12 / 0.31 $\pm$ 0.09 &  0.41 $\pm$ 0.14 / 0.43 $\pm$ 0.11 &   0.04 /  0.11  \\
    \multirow{2}{*}{(transformer encoders / }  & CTSR               &  0.32 $\pm$ 0.11 / 0.30 $\pm$ 0.11 & 0.43 $\pm$ 0.16 / 0.43 $\pm$ 0.15 &   0.06 /  0.09  \\
    \multirow{2}{*}{conv-only encoders)}      & TSR                 &  0.30 $\pm$ 0.11 / 0.24 $\pm$ 0.076 & 0.41 $\pm$ 0.14 / 0.40 $\pm$ 0.12 &  0.05 / 0.04  \\
                                           & only target sensor    &  0.26 $\pm$ 0.11 / 0.20 $\pm$ 0.09 & 0.38 $\pm$ 0.16 / 0.34 $\pm$ 0.13 &         n/a \\
    
    \midrule
    
     \multirow{2}{*}{Baseline all sensors}  &MC-CNN / \textbf{DeepConvLSTM}&  0.62 $\pm$ 0.11 / \textbf{0.63 $\pm$ 0.10} & 0.66 $\pm$ 0.9 / \textbf{ 0.66 $\pm$ 0.05} \\
                &Self-Attention / ST-Transformer& 0.55 $\pm$ 0.20 / 0.62 $\pm$ 0.08 & 0.62 $\pm$ 0.22 / 0.64 $\pm$ 0.11 \\
    \bottomrule
    \end{tabular}}
    \vspace{-15pt}
    \end{table*}
We evaluate our approach on two widely used HAR benchmark data sets: PAMAP2 \cite{reiss2012introducing} and OPPORTUNITY \cite{chavarriaga2013opportunity}. The PAMAP2 data set uses three wearable IMUs (on the chest, ankle, and wrist) to recognize 12 physical activities such as lying, sitting, standing, walking, running, cycling, Nordic walking, watching TV, computer work, driving, and ascending/descending stairs. OPPORTUNITY is a highly multimodal data set with tens of wearable and environmental sensors which also include chest, ankle, and wrist IMUs to recognize a variety of activities of daily life (ADLs) from a simulated morning routine (we evaluate it on the modes of locomotion classes). We focus on the three sensor locations from PAMAP2 and systematically investigate the benefit of knowledge transfer from one to the other. For both datasets we use a window size of 2s, with 0.5s slide in PAMAP2 and 1s slide in OPPORTUNITY.
\paragraph{Training Modes}
Regarding training we experimented with the following strategies:
	\begin{itemize}
		\item Training only with both modalities together, using only the classification loss. Of course, the testing is still only with the target modality. We name this strategy Training on a Shared Representation (TSR).
		\item Training with both modalities using classification loss on a shared representation like in (TSR), but this time also including the contrastive loss. We name this one Contrastive Together with a Shared Representation (CTSR).
		\item Pre-training first with only the contrastive loss for some epochs and then training with only the classification loss in the shared representation using both modalities.
		We denote Contrastive First, then Shared Representation (CFSR). 
		
		\item Training with only one modality, using only part of the classification loss. We will name those directly by their modalities (e.g. only Sensors). This is our baseline to which we compare.
	\end{itemize}
    All training was done using the Adam optimizer with a base learning rate of $0.001$ and $0.0001$ weight decay. The learning rate was reduced by a factor of $10$ every $100$ epochs. The temperature $\tau$ parameter used was $0.07$ and the batch size selected was $128$. If the classification loss is involved, training is done for $500$ epochs and the best model is selected using the validation set, always evaluated using only the target modality. If contrastive pre-training is performed, it is done for $100$ epochs with the same optimizer configurations.
    \paragraph{Results Discussion}
    The results of the experiments are shown in tables \ref{table:resPosEncodingDrillBetweenPositions} and \ref{table:res2}. When considering them it is important to note that the standard published  PAMAP2 and OPPORTUNITY data sets results (see bottom of table \ref{table:res2}) refer to full (or at least large) sensor sets. Especially in the OPPORTUNITY data set that is geared towards multi-modality, this means that the results we show have to be below the results with full sensor sets, as a single sensor contains significantly less information. 
    
    First and foremost we see that the use of our method leads to an across the board improvement of between 4\% and 11\% (on F1) when considering the average over all combinations of source and target sensors. While even just the shared representation training (TSR) provides a benefit, the contrastive approach is mostly better. At this stage, it is not possible to say if CFSR or CTSR is better with each slightly outperforming the other in different settings.
    
    In Table \ref{table:resPosEncodingDrillBetweenPositions} we give a more detailed breakdown of the improvement for the PAMAP2 case listing the individual sensor combinations and the best/worst performing sensors. While a detailed analysis is the subject of future work, there are some interesting observations. 
    For one we see that for certain sensor /activity combinations the gain can be much larger than the overall average (e.g. 40\% on the ankle to hand for descending stairs) while the worse combinations mostly simply have 0 gain (there is just one case of performance worsening). This is very encouraging for real-life applications.  The combinations where we see large gain are mostly in line with the intuition that the knowledge transfer mostly involves one sensor "nudging" the other to focus its representation on features that are relevant to the specific activity. Thus, as discussed in the introduction the problem with wrist-mounted sensors is that they contain signals from spurious motions (e.g. gesticulating)  that are much stronger than signals from mods of locomotion propagated through many joints. Thus it makes sense that both the chest and ankle significantly enhance the hand-mounted sensor performance on selected modes of locomotion. However, there are also cases that seem hard to understand and need further investigation such as the ankle helping the chest to recognize watching TV.

\section{Conclusions}
\label{sec:conclusions}
    
    The main conclusion of this work is that when facing an application that has to rely on a non-optimal sensor position using the proposed method to transfer knowledge from other sensor locations is a very promising approach. Not surprisingly it only works if the other sensor locations actually contain additional information and will have different effects on different activities depending on how much of the information refers to a specific activity. The method works for cases where the target sensors have very poor performance to start with as well as in cases where the performance is reasonable to start with. 

    As encouraging as the results are a number of questions remain for future work. First of all, is the performance when considering source and target sensors that differ in more than placement. We have performed initial experiments with motion data captured from video pose estimation and we hypothesize that as long as the sensors provide information that is related to the same physical aspects of the activity the method should be viable, but this needs to be systematically studied. Along the same lines is the question of using not just one, but multiple sensors as sources. On the methodological level more attention needs to be given to exploring various training modes e.g. iterating between contrastive and classification training, different batch sizing, and using adversarial training methods. Finally, we need to develop a deeper understanding of why and how the method works for different types of information.

\clearpage
\section*{Acknowledgements}
The research reported in this paper was supported by the BMBF (German Federal Ministry of Education and Research) in the project VidGenSense (01IW21003). It was also funded by the Carl Zeiss Stiftung under the Sunstaibable Embedded AI project (P2021-02-009).

\bibliographystyle{ACM-Reference-Format}
\bibliography{sample-base}

\end{document}